\documentclass[journal]{IEEEtran}
\usepackage{amssymb}
\usepackage[linesnumbered,ruled,vlined]{algorithm2e}
\SetKwInput{KwInput}{Input}                
\SetKwInput{KwOutput}{Output}              
\usepackage{mathrsfs}
\usepackage{multirow}
\usepackage{float}
\usepackage{amssymb}
\usepackage{subfigure} 
\usepackage[utf8]{inputenc}
\usepackage[small]{caption}
\usepackage{graphicx}
\usepackage{amsmath}
\usepackage{amsthm}
\usepackage{booktabs}
\usepackage{epstopdf}
\usepackage{pifont}
\usepackage{soul}
\usepackage{url}
\usepackage{color}

\usepackage{hyperref}
\hypersetup{
    colorlinks=true,
    linkcolor=red,
    filecolor=magenta,      
    urlcolor=cyan,
}

\usepackage{cite}

\ifCLASSINFOpdf
\else
\fi
\begin{document}
%
\title{MDFM: Multi-Decision Fusing Model for Few-Shot Learning}
%
%
%

\author{Shuai~Shao$^\dagger$,
        Lei~Xing$^\dagger$,
        Rui~Xu,
        Weifeng~Liu,~\IEEEmembership{Senior Member,~IEEE,}
        Yan-Jiang~Wang$^*$,
        Bao-Di~Liu$^*$,~\IEEEmembership{Member,~IEEE}
\thanks{Shao~Shuai, Rui~Xu, Weifeng~Liu, Yan-Jiang~Wang, and Bao-Di~Liu are with the College of Control Science and Engineering, China University of Petroleum (East China), 266580, China. 
Lei~Xing is with the College of Oceanography and Space Informatics, China University of Petroleum (East China), 266580, China.
}
\thanks{$^\dagger$Shao~Shuai and Lei~Xing are co-first authors. }
\thanks{$^*$Bao-Di~Liu (Email: thu.liubaodi@gmail.com) and Yan-Jiang~Wang (Email: yjwang@upc.edu.cn) are corresponding authors.}}

%
%

\markboth{Journal of \LaTeX\ Class Files,~Vol.~14, No.~8, August~2015}%
{Shell \MakeLowercase{\textit{et al.}}: Bare Demo of IEEEtran.cls for IEEE Journals}
%



\maketitle

\begin{abstract}
In recent years, researchers pay growing attention to the few-shot learning (FSL) task to address the data-scarce problem.
A standard FSL framework is composed of two components: i) Pre-train. Employ the base data to generate a CNN-based feature extraction model (FEM). ii) Meta-test. Apply the trained FEM to the novel data (category is different from base data) to acquire the feature embeddings and recognize them.
Although researchers have made remarkable breakthroughs in FSL, there still exists a fundamental problem.
Since the trained FEM with base data usually cannot adapt to the novel class flawlessly, the novel data's feature may lead to the distribution shift problem. 
To address this challenge, we hypothesize that even if most of the decisions based on different FEMs are viewed as \textit{weak decisions}, which are not available for all classes, they still perform decent in some specific categories.
Inspired by this assumption, we propose a novel method \textbf{Multi-Decision Fusing Model (MDFM)}, which comprehensively considers the decisions based on multiple FEMs to enhance the efficacy and robustness of the model.
MDFM is a simple, flexible, non-parametric method that can directly apply to the existing FEMs.
Besides, we extend the proposed MDFM to two FSL settings (e.g., supervised and semi-supervised settings).
We evaluate the proposed method on five benchmark datasets and achieve significant improvements of \textbf{3.4\%}-\textbf{7.3\%} compared with state-of-the-arts.
\end{abstract}

\begin{IEEEkeywords}
Few-Shot Learning (FSL), distribution shift problem, Multi-Decision Fusing Model (MDFM)
\end{IEEEkeywords}

%
\IEEEpeerreviewmaketitle

\section{Introduction}


In recent years, deep learning, as a powerful tool, has helped machines reached or even surpass human beings' level in various visual tasks, such as image classification \cite{wang2016cost,shao2020label,yao2020deep}, person re-identification \cite{wang2020dense,zheng2018pedestrian,fan2020contextual}, 
blind image quality assessment \cite{ma2017end,hosu2020koniq,zhang2021uncertainty}, vision-and-language navigation \cite{anderson2018vision,zhang2020language,majumdar2020improving}.
One indispensable factor is attributed to the large-scale labeled data.
However, as the limitation of actual circumstances, it may be infeasible to collect large amounts of labeled data in the real world. 
Thus, few-shot learning (FSL), targets to address this problem with scarce labeled samples, has attracted growing attention.
Generally, the current popular FSL model usually includes two components: i) Pre-train. Employ the base data $\mathcal{D}_{base}$ to generate a CNN-based feature extraction model (FEM). 
ii) Meta-test. First, extract the feature embeddings of novel data $\mathcal{D}_{novel} = \{\mathcal{S}, \mathcal{U}, \mathcal{Q}\}$, where $\mathcal{S}$, $\mathcal{U}$ and $\mathcal{Q}$ denote support set, unlabeled set and query set. Next, design a classifier to recognize the query samples. 
For more details, please refer to Section \ref{sec: Problem Formulation}.


\begin{figure}[t]
	\begin{center}
		\includegraphics[width=1.0\linewidth]{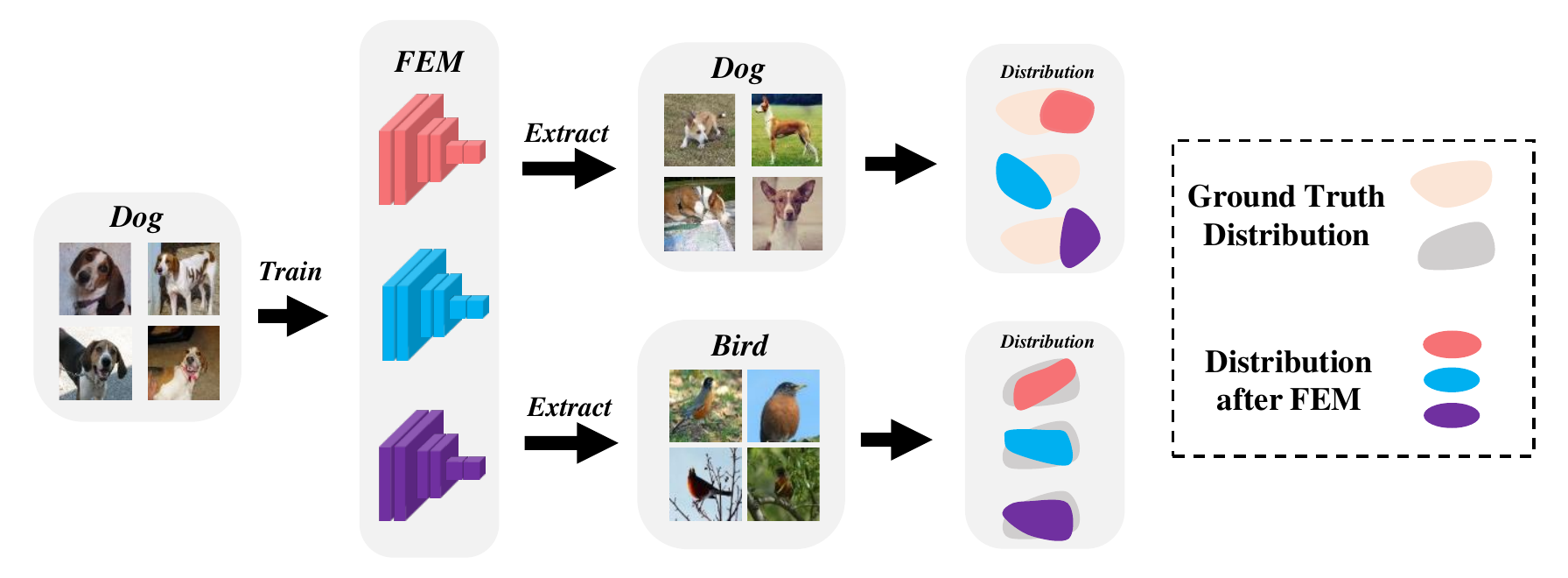}
	\end{center}
	\caption{
    An example to introduce distribution shift problem. 
    The feature extraction model (FEMs) are trained with some dog images.
    We can achieve appropriate feature distributions when using these FEMs to extract other dog images but deviated feature distributions on extracting bird images.}
	\label{fig: example}
\end{figure}

\begin{figure*}
	\begin{center}
		\includegraphics[width=1.0\linewidth]{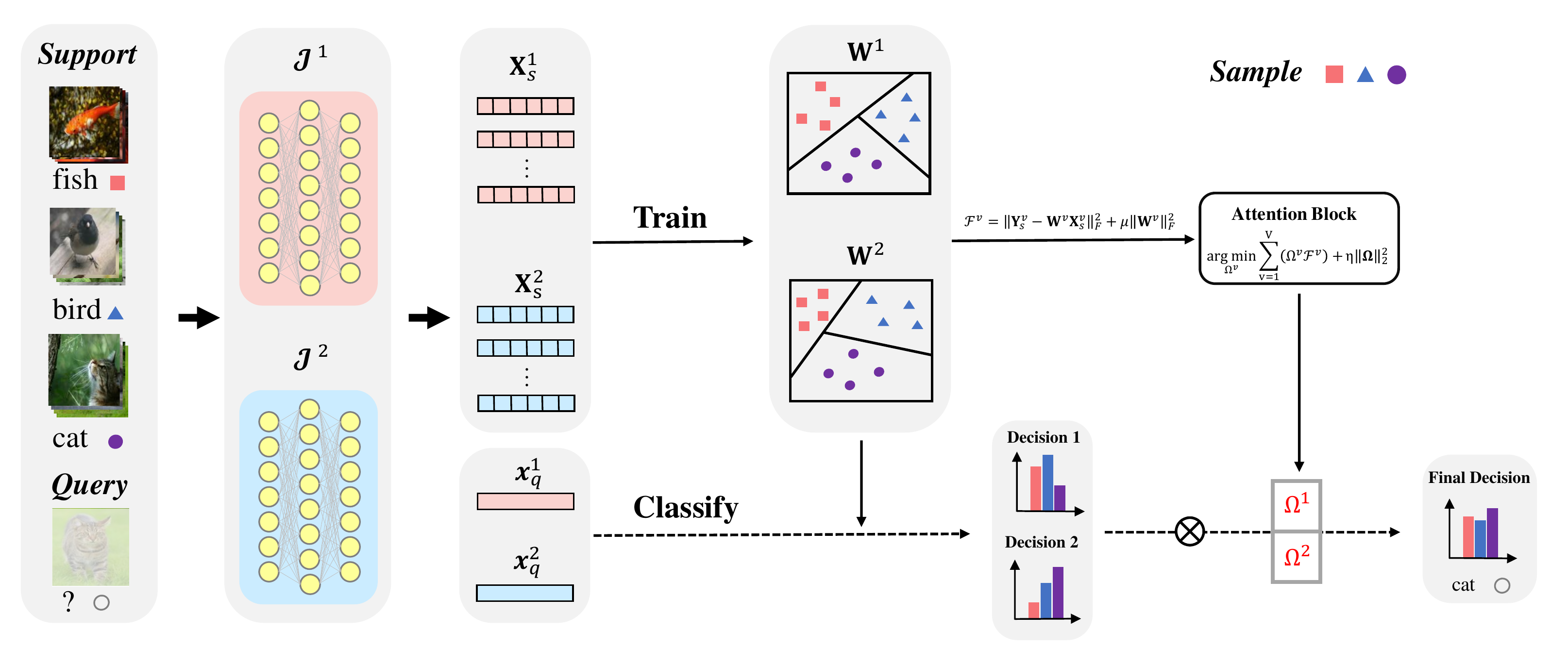}
	\end{center}
	\caption{
	The complete framework of our Multi-Decision Fusing Model (MDFM) on supervised setting. 
	Assume we have two views of feature extraction models (FEMs), e.g., $\mathcal{J}^v$, where $v=[1,2]$ denotes the $v_{th}$ view. There are a total of $4$ steps. 
	\textbf{(1)} Input images to FEMs and obtain the support feature $\mathbf{X}_s^v$ and query feature $\mathbf{x}_q^v$.
	\textbf{(2)} Exploit support feature $\mathbf{X}_s^v$ to train classifiers $\mathbf{W}^v$.
	\textbf{(3)} Learn the combination weights $\mathbf{\Omega}$ for each view.
	\textbf{(4)} Use $\mathbf{W}^v$ to classify query data and achieve multi-decisions (use probability charts to represent).
	\textbf{(5)} Assert ${\Omega}^v$ to the corresponding decision and fuse them for final prediction.}
	\label{fig: Flowcharts}
\end{figure*}

To improve the performance of FSL, most works pay attention on designing a more robust and powerful FEM by introducing lots of strategies, such as self-supervised learning \cite{rodriguez2020embedding}\cite{mangla2020charting}, meta-learning \cite{snell2017prototypical}\cite{lee2019meta}, graph structure \cite{liu2019learning}\cite{yang2020dpgn}, knowledge distillation \cite{dvornik2019diversity}\cite{rizve2021exploring}.
Actually, these FEMs (trained from the base set) perform well when extracting base features, which makes the distribution based on any FEM close to the ground truth sample distribution. Just like the \textit{dogs' distribution} in Figure \ref{fig: example}.
However, there is a fundamental problem in FSL: no matter how well the FEMs perform on the base class, they can not adapt to the novel class flawlessly due to the cross-domain limitation.
Therefore, the novel samples' distributions based on the FEMs usually have a certain degree of deviation compared with the ground truth distributions, and they are very different from each other. An example of \textit{birds' distribution} is illustrated in Figure \ref{fig: example}. 
This is a typical distribution shift problem in transfer learning and domain adaption.

To address this issue, it sounds like we only need to fine-tune the network structure to accommodate the new class. However, as the scarce of labeled novel data (as an example, on typically $1$-shot or $5$-shot case, each category only has $1$ or $5$ labeled sample), this kind of method performs poorly on FSL, which has been proved in MAML \cite{finn2017model}. 
Following, \cite{dvornik2019diversity} proposed ensemble method to fuse multiple designed networks to tackle distribution shift problem. 
But this approach relies on the specialized FEM in the pre-train phase and specific classifier in the meta-test phase and only has a limited promotion for FSL.
To this end, dedicated technology is necessary.

In this paper, we propose a 
Multi-Decision Fusing Model (MDFM)
to solve this challenge from the perspective of fusing multi-decision.
Specifically, assume we have multiple views. Each view corresponds to a strategy to design the FEM in the pre-train stage (for example, the first view's FEM follows MetaOptNet \cite{lee2019meta}, and the second follows ICINet \cite{wang2020instance}).
From Figure \ref{fig: example}, we find that the bird samples' features distribute variants on different views, which means we can achieve multiple kinds of decisions through these features.
Despite all the decisions are viewed as the \textit{weak decisions} due to distribution shift problem, which may be not available for all classes, they still
perform decent 
in some specific categories (we have evaluated this conclusion through confusion matrix in Figure \ref{fig: ablation_studie_2}). 
Therefore, we attempt to select the proper \textit{weak decisions} for each class and then fuse all classes' decisions. 

To achieve this purpose, we consider the training loss as the criterion and design a weighting mechanism (more details, please refer to Section \ref{sec: Methodology}). This way helps the \textit{weak decisions} positively impact the final results.
Different from \cite{dvornik2019diversity}, our MDFM is a simple, flexible, non-parametric model that can directly apply to all kinds of FEMs and classifiers.
Besides, according to the data adopted in the design of the classifier, researchers categorize the FSL-based approaches as two sorts: 
i) Supervised Few-Shot Learning (SFSL), ii) Semi-Supervised Few-Shot Learning (SSFSL). 
The difference is that SSFSL uses unlabeled training data, while SFSL does not. In this paper, we extend our MDFM to the two settings.
For convenience, we list some crucial abbreviations and notations in Table \ref{table: definition}, and illustrate the overall flowchart of MDFM in Figure \ref{fig: Flowcharts}.

In summary, the main contributions focus on:
\begin{itemize}
\item 
We propose a novel method for FSL, dubbed as Multi-Decision Fusing Model (MDFM). It comprehensively considers the \textit{weak decisions} on multiple views to increase the efficacy and robustness of the FSL framework and solve the distribution shift problem.

\item
Compared with the existed ensemble method \cite{dvornik2019diversity}, MDFM is a simple, flexible, non-parametric method that is not restricted by the FEM and classifier. 
This highlight gives us more options to update the components on the FSL framework to improve classification accuracy, which may help apply the FSL framework in reality.

\item
We evaluate the proposed method on four benchmark datasets (mini-ImageNet, tiered-ImageNet, CIFAR-FS, FC100) and achieve significant improvements of 1.9\%-7.1\% compared with other state-of-the-art methods. Besides, to evaluate the robustness of the proposed method, we design the cross-domain experiments on the CUB dataset and achieve far better performance than state-of-the-art methods of at least 7.9\%.
\end{itemize}

\begin{table*}[t]
        \caption{Some important abbreviations and notations.}
        \begin{center}
        \setlength{\tabcolsep}{10mm}{
            \begin{tabular}{ll}
                \toprule
                \textbf{Abbreviation and Notation}                  & \textbf{Definition}      \\
                \midrule

                FSL          & few-shot learning \\
                FEM          & feature extraction model \\
                MDFM         & multi-decision fusing model \\
                
                Std-Dec      & standard decision \\
                Meta-Dec     & meta decision \\
                SS-R-Dec     & self-supervised rotation decision \\
                SS-M-Dec     & self-supervised mirror decision \\

                \midrule
                $\mathcal{D}_{base}$, $\mathcal{D}_{novel}$  & base data, novel data \\
                $\mathcal{S}$, $\mathcal{Q}$, $\mathcal{U}$ & support set, query set, unlabeled set \\
                $\mathcal{J}^v(\cdot)$ & CNN-based FEM on the $v_{th}$ view \\
                
                $\mathbf{X}^v$, $\mathbf{x}^v_{ts}$ & features of training and testing data on the $v_{th}$ view \\
                $\mathbf{X}_s^v$, $\mathbf{X}_u^v$, $\mathbf{X}_q^v$ & features of support, unlabeled, query data on the $v_{th}$ view\\
                $\mathbf{x}_{select}^v$, $\mathbf{y}_{select}^v$ & most confident sample feature and corresponding label on the $v_{th}$ view in self-training process\\ 
                $\mathbf{Y}$ & label matrices of training data \\
                $\mathbf{Y}_s^v$, $\mathbf{Y}_u^v$  & label matrices of support, unlabeled data\\  
                $\mathbf{W}^v$ & classifier on the $v_{th}$ view\\
                $\mathbf{\Omega} = [\Omega^1,\Omega^2,\dots,\Omega^V]^T$ & combination weights for different decisions\\
              \bottomrule
            \end{tabular} 
        }
        \end{center}
        \label{table: definition}        
\end{table*}

\section{Related Work}
\label{sec: Related Work}
\subsection{Few-Shot Learning}
In the past decade, FSL based works have attracted lots of attention. Researchers have proposed various classical frameworks to solve this problem. We list the two most popular types, including 
i) Meta-learning based methods, such as MAML \cite{finn2017model}, Reptile \cite{nichol2018first}, LEO \cite{rusu2019meta}, which purpose to obtain a universal model to rapidly adapt to new tasks. 
ii) Metric learning based methods, focusing on looking for ideal distance metrics to strengthen model's robustness, including ProtoNet \cite{snell2017prototypical}, MetaOpt \cite{lee2019meta}, TADAM \cite{oreshkin2018tadam}, MSML \cite{jiang2020multi} $et. al.$.

In addition, all these methods can be split into another taxonomy, e.g. supervised few-shot learning (SFSL), and semi-supervised few-shot learning (SSFSL). For example, MAML \cite{finn2017model}, LEO \cite{rusu2019meta}, S2M2 \cite{mangla2020charting}, DPGN \cite{yang2020dpgn}, TEAM \cite{qiao2019transductive}, SIB \cite{hu2020empirical}, IPBT \cite{zhang2019few} $et.al.$ are based on supervised setting, only use the labeled support data to train the classifier; and LST \cite{li2019learning}, EPNet \cite{rodriguez2020embedding}, ICI \cite{wang2020instance}, MHFC \cite{shao2021mhfc} $et.al.$ are based on semi-supervised setting, employ both support and unlabeled data to train the classifier. 
Our method is applicable to both the two settings and achieves outstanding performance.

\subsection{Distribution Shift Problem in FSL}
Distribution shift problem is a typical problem in many fields, such as transfer learning, domain adaption, domain generalization, which also exist in the FSL.
In FSL, researchers usually address this problem from two perspectives. i) On the one hand, researchers design more robust FEM, make it adapt to the novel, unseen classes, such as introducing self-supervised learning \cite{mangla2020charting} and meta-learning \cite{rodriguez2020embedding} strategies .
ii) On the other hand, researchers fix the FEM and pay attention on processing the extracted feature embeddings, make them more discriminative, such as the distribution calibration \cite{yang2021free} and instance credibility inference \cite{wang2020instance}.



\subsection{Multi-View Learning}
Just as every coin has two sides, it would be incomplete to define objects from a single perspective. Therefore, multi-view learning has received wide attention in recent years.
There exist lots of classical methods and corresponding applications. For example, 
Liu $et. al.$ proposed a sparse coding based multi-view method MHDSC \cite{liu2014multiview} for image annotation task; 
Liu $et. al.$ proposed SPM-CRC \cite{liu2019weighted}, which improve the collaborative representation model from multi-view learning to classify remote sensing images; 
Jan~$et. al.$ proposed MVCCA~\cite{rupnik2010multi} and employed it in natural language processing.
Liu $et. al.$ proposed MHL \cite{liu2017multi} to solve Alzheimer’s Disease Predicting problem; 
Zhang~$et. al.$ proposed IMHL~\cite{zhang2018inductive}, which is an inductive hypergraph learning from multi-view and applied it for $3D$ object recognition.
All these methods may help FSL, and some multi-view based works have been proposed.

DenseCls \cite{lifchitz2019dense} splits the feature map into different blocks and predicts the corresponding label.
DivCoop \cite{dvornik2020selecting} employs various datasets to train the FEMs and fuse them to a multi-domain representation. 
DWC \cite{dvornik2019diversity} design an ensemble model with a cooperation strategy to fuse multiple information.
URT \cite{liu2021universal} is the improvement of DivCoop \cite{dvornik2020selecting}, which introduces a transformer layer to better employing the different datasets.
Just like our MDFM, all of these methods are based on multi-view learning. But they are restricted by the fixed FEMs and classifiers, which lose scalability.
This paper, inspired by the traditional multi-view method and ensemble learning strategy, mainly focuses on designing a novel, flexible ensemble-based multi-view framework to address distribution shift problem and extending it to two FSL settings. 



\section{Problem Formulation}
\label{sec: Problem Formulation}
In this section, we focus on introducing the few-shot learning model. Two components, such as pre-train and meta-test, are involved in the FSL procedure. 
In pre-train phase, we assume that $\mathcal{D}_{base} = \{(x_i,y_i) |{\kern 1pt} y_i \in \mathcal{C}_{base} \}_{i=1}^{N_{base}}$ represents the base dataset, where $\mathcal{C}_{base}$ denotes the base category set, $x$ and $y$ indicate the sample and corresponding label, respectively. $N_{base}$ denotes the total number of base data.
We train the CNN-based FEM $\mathcal{J}(\cdot)$ on $\mathcal{D}_{base}$.
In this paper, we design several kinds of FEMs from different views, and define the FEM on the $v_{th}$ view as $\mathcal{J}^v(\cdot)$, where $v=1,2,\cdots,V$. More details please refer to Section \ref{subsec: Discussion about Multiple Decisions}.

Next to the meta-test phase, utilising the $\mathcal{J}^v(\cdot)$ to extract features for novel dataset $\mathcal{D}_{novel} = \{(x_j,y_j) |{\kern 1pt} y_j \in \mathcal{C}_{novel} \}_{j=1}^{N_{novel}}$, where $\mathcal{C}_{novel}$ denotes the novel category set, $C_{base} {\kern 2pt} \cap {\kern 2pt} C_{novel} = \emptyset$. $N_{novel}$ denotes the number of novel data.
Besides, the novel dataset composed of three components, e.g., $\mathcal{D}_{novel} = \{\mathcal{S}, \mathcal{U}, \mathcal{Q}\}$, where $\mathcal{S}$, $\mathcal{U}$ and $\mathcal{Q}$ denote support set, unlabeled set and query set, 
$\mathcal{S} {\kern 2pt} \cap {\kern 2pt} \mathcal{U}  = \emptyset$, 
$\mathcal{S} {\kern 2pt} \cap {\kern 2pt} \mathcal{Q}  = \emptyset$,
$\mathcal{Q} {\kern 2pt} \cap {\kern 2pt} \mathcal{U}  = \emptyset$.
Finally, we design a classifier to classify $\mathcal{Q}$. 
The to-be-designed classifier includes two settings, e.g., supervised setting, semi-supervised setting. Section \ref{subsec: Supervised MDFM}, \ref{subsec: Semi-Supervised MDFM} show more details.
We follow standard $C$-$way$-$M$-$shot$ per episode as \cite{wang2020instance} for classification, where $C$-$way$ denotes $C$ classes, and $M$-$shot$ indicates $M$ samples per class. We average the accuracies of all the episodes with $95\%$ confidence intervals as the final result.

\section{Methodology}
\label{sec: Methodology}
In this section, first, we propose a Multi-Decision Fusing Model (MDFM) for few-shot learning and show the details in Section \ref{subsec: Multi-Decision Fusing Model}. Then, we extend MDFM to varied settings (e.g., supervised and semi-supervised settings) in Section \ref{subsec: Supervised MDFM}, \ref{subsec: Semi-Supervised MDFM}. 
Next, we discuss the details about multiple decisions and define the corresponding FEMs in Section \ref{subsec: Discussion about Multiple Decisions}.
Finally, we analyse the complex in Section \ref{subsec: complex}.

\subsection{Multi-Decision Fusing Model}
\label{subsec: Multi-Decision Fusing Model}
It is worth noting that the proposed model can integrate all types of conventional classification strategies (such as support vector machine, logistic regression, linear regression). In this paper, we merely consider a simple linear regression model as an example. The objective function of the linear regression classifier is as follows:
\begin{equation}
\begin{split}
        \mathop {\arg \min}\limits_{\mathbf{W}} \mathcal{F}
        = \left\| \mathbf{Y} - \mathbf{W} \mathbf{X} \right\|_F^2
        + \mu \left \| \mathbf{W} \right\|_F^2
\end{split}
\label{eqa: linear_regression_obj}
\end{equation}
where 
$\mathbf{X}=[{\mathbf{x}}_1,{\mathbf{x}}_2,\dots,{\mathbf{x}}_{N}] \in \mathbb{R}^{dim \times N}$, $\mathbf{Y}=[{\mathbf{y}}_1,{\mathbf{y}}_2,\dots,{\mathbf{y}}_{N}] \in \mathbb{R} ^{C \times N}$, $dim$ and $N$ indicate the dimension and number of labeled samples, respectively. $C$ denotes the number of categories.
${\mathbf{x}}_i,{\mathbf{y}}_i$ ($i = 1, 2, \dots$) denote the feature embedding vector and one-hot label vector of the $i_{th}$ sample.
$\mathbf{W} \in \mathbb{R} ^{C \times dim}$ represents the to-be-learned classifier. 
We directly optimize the objective function and obtain the $\mathbf{W}$ as:
\begin{equation}
\begin{split}
        \mathbf{W} = \mathbf{Y}{\mathbf{X}}^T \left(\mathbf{X}{\mathbf{X}}^T+\mu \mathbf{I} \right)^{-1}
\end{split}
\label{eqa: single_view_classifier}
\end{equation}

Following, given a testing sample feature $\mathbf{x}_{ts} \in \mathbb{R}^{dim \times 1}$, we predict the $\mathbf{x}_{ts}$'s category by:
\begin{equation}
\begin{split}
        \mathcal{C}(\mathbf{x}_{ts}) = onehot \left\{
        id_{max}
        \left\{
        \mathbf{W} \mathbf{x}_{ts}  
        \right\}\right\}
\end{split}
\label{eqa: single_view_predict}
\end{equation}
where $id_{max}$ denotes an operator to obtain the index of the max value in the vector. $onehot$ indicates the operator to generate a one-hot label.

To sufficiently extract more information of few-shot data in real applications, we introduce multiple feature representations for samples from different views.
Assume that we have $V$ views in total, each view has the corresponding feature embedding and classifier, e.g., $\left[(\mathbf{X}^1,\mathbf{W}^1), (\mathbf{X}^2,\mathbf{W}^2), \dots, (\mathbf{X}^V,\mathbf{W}^V) \right]$, where $(\cdot)^v,(v=1,2,\dots,V)$ denotes the variable on the $v_{th}$ view.
And each view obtains a decision by using Equation (\ref{eqa: single_view_predict}). 
We try to find the combination weights $\mathbf{\Omega} = [\Omega^1,\Omega^2,\dots,\Omega^V]^T$ to make the weak classifiers have a positive impact on the final decision, the objective function is formulated as:
\begin{equation}
\begin{split}
        \mathop {\arg \min}\limits_{\mathbf{\Omega}} 
        \mathcal{G} =\sum_{v=1}^V 
        \left (
        \Omega^v \mathcal{F}^v 
        \right )
        + \eta \left\| \mathbf{\Omega} \right\|_2^2\\
		\text{s.t.} {\kern 4pt}  
		 \sum_{v=1}^V \Omega^v = 1, {\kern 4pt} \Omega^v \ge 0
\end{split}
\label{eqa: multi_view_classifier}
\end{equation}
where 
$\Omega^v$ indicates the weight of $v_{th}$ view.
$\mathcal{F}^v$ indicates the (Equation (\ref{eqa: linear_regression_obj}))'s loss of $v_{th}$ view. 
We introduce the Lagrangian to solve the problem, the Equation (\ref{eqa: multi_view_classifier}) is rewritten as:
\begin{equation}
\begin{split}
        \mathop {\arg \min}\limits_{\mathbf{\Omega, \zeta, \Lambda}} 
        \mathcal{G}=
        \sum_{v=1}^V 
        \left (
        \Omega^v \mathcal{F}^v 
        \right )
        + \eta \left\| \mathbf{\Omega} \right\|_2^2 \\
         - \zeta \left(\sum_{v=1}^V \Omega^v - 1 \right)
        -\mathbf{\Lambda}^T \mathbf{\Omega}
\end{split}
\label{eqa: obj_compute_Omega_Lagrangian}
\end{equation}
where $\zeta$ is a constant, $\mathbf{\Lambda} = [\Lambda^1,\Lambda^2,\dots,\Lambda^V]^T$ is a vector.
Assume $\hat{\mathbf{\Omega}}$, $\hat{\zeta}$, $\hat{\mathbf{\Lambda}}$ are the optimal solutions, we solved this problem as:
\begin{equation}
\begin{split}
        \hat{\Omega}^v = \frac{1}{2\eta} 
        max\left\{
        \frac{\sum_{v=1}^V \mathcal{F}^v}{V} + \frac{2\eta}{V} - \mathcal{F}^v - \hat{\Lambda}_{avg} ,0\right\}\\
\end{split}
\label{eqa: omega_final}
\end{equation}
where $\hat{\Lambda}_{avg}$ is a constant, denotes the average of $\hat{\mathbf{\Lambda}}$.
For the detailed optimization process, please refer to Supplementary Material.
Then, we employ the proposed MDFM to predict the testing sample $\mathbf{x}_{ts}$, the Equation (\ref{eqa: single_view_predict}) is rewritten as:
\begin{equation}
\begin{split}
        \mathcal{C}(\mathbf{x}_{ts}) = onehot \left\{
        id_{max} 
        \left\{
        \sum_{v=1}^V \Omega^v \mathbf{W}^v \mathbf{x}_{ts}^v  
        \right\}\right\}
\end{split}
\label{eqa: multi_view_predict}
\end{equation}
where $\mathbf{W}^v = \mathbf{Y}{\mathbf{X}^v}^T \left(\mathbf{X}^v{\mathbf{X}^v}^T+\mu \mathbf{I} \right)^{-1}$, $\mathbf{x}_{ts}^v$ is the feature embedding of $\mathbf{x}_{ts}$ on the $v_{th}$ view.

\subsection{Supervised MDFM}
\label{subsec: Supervised MDFM}
Define the feature embedding of $\mathcal{D}_{novel}$ on the $v_{th}$ view as ${\mathbf{X}_{novel}^v}=[{\mathbf{X}_s^v}, {\mathbf{X}_u^v}, {\mathbf{X}_q^v}]$, where ${\mathbf{X}_s^v}=\mathcal{J}^v(\mathcal{S})$, ${\mathbf{X}_u^v}=\mathcal{J}^v(\mathcal{U})$, and ${\mathbf{X}_q^v}=\mathcal{J}^v(\mathcal{Q})$ denote the feature embeddings of support, unlabeled, and query data on the $v_{th}$ view.
Researchers employ different data to design the classifier, and these methods can be split into two settings, e.g., supervised setting and semi-supervised setting.

Supervised setting in few-shot learning adopt the support set $\mathcal{S}$ to train the classifier and directly predict the query set $\mathcal{Q}$'s category.
We directly utilize the MDFM to achieve this purpose by:
\begin{equation}
\begin{split}
        \left\{\begin{array}{llll}
            \mathcal{F}^v
            = \left\| {\mathbf{Y}_s^v} - {\mathbf{W}}^v {\mathbf{X}_s^v} \right\|_F^2
            + \mu \left \| \mathbf{W}^v \right\|_F^2\\
            \mathbf{W}^v = \mathbf{Y}_s^v{\mathbf{X}_s^v}^T \left(\mathbf{X}_s^v{\mathbf{X}_s^v}^T+\mu \mathbf{I} \right)^{-1}\\        
            \hat{\Omega}^v = \frac{1}{2\eta}
            max\left\{
            \frac{\sum_{v=1}^V \mathcal{F}^v}{V} + \frac{2\eta}{V} - \mathcal{F}^v - \hat{\Lambda}_{avg} ,0\right\}\\
            \mathcal{C}(\mathbf{X}_q^v) = onehot \left\{ id_{max}
            \left\{
            \sum_{v=1}^V \Omega^v \mathbf{W}^v {\mathbf{X}_q^v}  
            \right\} \right\}         
        \end{array}\right.
\end{split}
\label{eqa: inductive_classifier}
\end{equation}
where $\mathbf{Y}_s^v$ denotes the one-hot label matrix of support data on the $v_{th}$ view. Note that, on the supervised setting, the ${\mathbf{Y}_s}$ of different views are the same. 


\subsection{Semi-Supervised MDFM}
\label{subsec: Semi-Supervised MDFM}
Unlike supervised few shot learning, on semi-supervised setting, besides the support data's feature embeddings and label information, researchers also apply the feature embeddings of unlabeled data to construct the classifier and then predict the query label. 
This paper extends MDFM to semi-supervised setting by introducing a simple self-training strategy to strengthen the classifier. We show the detailed steps as:

\textbf{i)}
Train a basic classifier by employing the support data $\mathcal{S}$, and then utilize the trained classifier to predict the unlabeled data $\mathcal{U}$ by:
\begin{equation}
\begin{split}
        \left\{\begin{array}{ll}
            \mathbf{W}^v = \mathbf{Y}_s^v{\mathbf{X}_s^v}^T \left(\mathbf{X}_s^v{\mathbf{X}_s^v}^T+\mu \mathbf{I} \right)^{-1}\\        
            {\mathbf{Y}_u^v} = \mathbf{W}^v {\mathbf{X}_u^v}
        \end{array}\right.
\end{split}
\label{eqa: predict_unlabel_data}
\end{equation}
where ${\mathbf{Y}_u^v}$ denotes the predicted soft label matrix of unlabeled data on the $v_{th}$ view.

\textbf{ii)}
Follow traditional self-training strategy \cite{raina2007self},
select one most confident sample $\mathbf{x}_{select}^v$ through the ${\mathbf{Y}_u^v}$ without putting back, the corresponding one-hot pseudo label is denoted as $\mathbf{y}_{select}^v$. Then, expand it to the support data by:
\begin{equation}
\begin{split}
        \left\{\begin{array}{ll}
            {\mathbf{X}_s^v} 
            = \left[{\mathbf{X}_s^v},\mathbf{x}_{select}^v  \right]\\
            {\mathbf{Y}_s^v} 
            = \left[{\mathbf{Y}_s^v},\mathbf{y}_{select}^v  \right]
        \end{array}\right.
\end{split}
\label{eqa: select_most_confidence_unlabel_data}
\end{equation}

\textbf{iii)}
Repeat \textbf{i)} and \textbf{ii)} until the performance of classifiers are stable. Finally, employ the optimal classifiers on different views to predict the query label by Equation (\ref{eqa: inductive_classifier}). Note that, when we start updating the basic classifier, the label matrices $\mathbf{Y}_s^v,(v=1,2,\dots)$ would be different on different views.
We summarize the steps in Algorithm \ref{Algorithm: SS-MDFM}.

\begin{algorithm}[t]
\DontPrintSemicolon
  
    \KwInput{Base set $\mathcal{D}_{base}$, Novel set $\mathcal{D}_{novel}=\{\mathcal{S},\mathcal{U},\mathcal{Q}\}$}
    \KwOutput{Query label}
    
    Design the multi-view feature extraction model 
    $\mathcal{J}^v(\cdot)$, and obtain feature embeddings by $\mathbf{X}_s^v=\mathcal{J}^v(\mathcal{S})$, $\mathbf{X}_u^v=\mathcal{J}^v(\mathcal{U})$, $\mathbf{X}_q^v=\mathcal{J}^v(\mathcal{Q})$.\\

    \Repeat{the performance of to-be-learned classifiers are stable.}
    {
    Train a basic classifier $\mathbf{W}^v$ through  $\mathbf{X}_s^v$, and use it to predict the unlabeled data by Equation (\ref{eqa: predict_unlabel_data})\\
    Select the most confident sample and expand it to the support set by Equation (\ref{eqa: select_most_confidence_unlabel_data}).\\
    }
    Utilize the optimal classifier to predict the query label by Equation (\ref{eqa: inductive_classifier}).
    
	\caption{Semi-Supervised MDFM}
	\label{Algorithm: SS-MDFM}
\end{algorithm}

\subsection{Discussion about Multiple Decisions}
\label{subsec: Discussion about Multiple Decisions}
The to-be-fused decisions are determined by the corresponding feature extraction models (FEMs), which have a large number of choices. 
As examples:
\textbf{(1)}
Standard decision (Std-Dec), the FEM utilizes a standard CNN-based classification structure, such as \cite{wang2020instance}.
\textbf{(2)}
Meta decision (Meta-Dec), the FEM introduces the meta-learning strategy to the network, just like \cite{lee2019meta}.
\textbf{(3)}
Self-supervised decision (SS-Dec), the FEM adds auxiliary losses to the standard CNN-based classification structure from a self-supervised perspective to strengthen the robustness of the network, similar as \cite{mangla2020charting}.
We show the results of all kinds of fusing ways in Table \ref{tab: Multi_Feature_Fusion}.

In this paper, most experimental results are merely based on two kinds of SS-Decs.
For the first category, we design the FEM by introducing standard classification loss $\mathcal{L}_c$ to predict the sample labels, and auxiliary rotation loss $\mathcal{L}_r$ (e.g., rotate the dataset to $r$ degree and $r \in \mathcal{C_R} = \{0^{\circ}, 90^{\circ}, 180^{\circ}, 270^{\circ} \}$) to predict image rotations. 
The decision based on this kind of FEM is dubbed as SS-R-Decision (SS-R-Dec). The first loss function is defined as $\mathcal{L}_c + \mathcal{L}_r$, we define $\mathcal{L}_c$ and $\mathcal{L}_r$ as:
\begin{equation}
\begin{split}
        \mathcal{L}_{c}
        =-\sum_{c}
         y_{(c,x)}log(p_{(c,x)})\\
\end{split}
\label{eqa: classification_loss}
\end{equation}
where $c \in \mathcal{C}_{base}$ denotes the $c_{th}$ class. $y_{(c,x)}$, $p_{(c,x)}$ indicate the probabilities that the truth label and predicted label of $x_{th}$ sample belongs to $c_{th}$ class.

\begin{equation}
\begin{split}
        \mathcal{L}_{r}
        =-\sum_{r}
         y_{(r,x)}log(p_{(r,x)})\\
\end{split}
\label{eqa: rotation_loss}
\end{equation}
where $y_{(r,x)}$, $p_{(r,x)}$ indicate the probabilities that the truth label and predicted label of $x_{th}$ sample belongs to $r_{th}$ class.

The second decision is SS-M-Decision (SS-M-Dec), which adopts another strategy to train the FEM. Specifically, we design the SS-M-Dec by adding the loss $\mathcal{L}_c$ and auxiliary mirror loss $\mathcal{L}_m$ (e.g., mirror the samples with $m$ ways and $m \in \mathcal{C_M} = \{vertically, horizontally, diagonally \}$) to the network to predict image mirrors.
We summarize the second loss function as $\mathcal{L}_c + \mathcal{L}_m$, and define the $\mathcal{L}_m$ as:
\begin{equation}
\begin{split}
        \mathcal{L}_{m}
        =-\sum_{m}
         y_{(m,x)}log(p_{(m,x)})\\
\end{split}
\label{eqa: mirror_loss}
\end{equation}
where $y_{(m,x)}$, $p_{(m,x)}$ indicate the probabilities that the truth label and predicted label of $x_{th}$ sample belongs to $m_{th}$ class.

\subsection{Complexity Analyse}
\label{subsec: complex}
In our method, as that the pre-trained network in each view is based on the ResNet-12 backbone, their floating point operations (FLOPs) are fixed. 
Next, we merely talk about the complexity in the meta-test stage. 
It can be predicted that the complexity of our multi-view fusion method is related to the number of views. Assume we have $4$ views. Our model takes about $4$ times as long to process a single image as a single-view model. But fortunately, on the Tesla-V100 GPU, the single-view model just spends $9$ milliseconds to process an image, and our method needs $36$ milliseconds. This kind of consumption is acceptable. Besides the improvement of classification accuracy, our method maybe a feasible way in reality.

\section{Experiments}
In this section, we first briefly review the benchmark datasets and show the implementation details. Then, we list the supervised experimental results in Table \ref{table: comparison_results_mini_tiered}, \ref{table: comparison_results_cifar_fc100}, and semi-supervised results in Table \ref{table: semi_comparison_results_mini_tiered}, then analyse them. Next, we perform ablation studies and discuss the multiple decisions to study what influences the performance.
In the following, we conduct a cross-domain experiment to further evaluate the ability and robustness of the proposed method.
We conduct all the experiments on a Tesla-$V100$ GPU with $32G$ memory. 

\begin{table*}[!t]
\caption{The $5$-way supervised few-shot classification accuracies on mini-ImageNet and tiered-ImageNet with $95\%$ confidence intervals over $600$ episodes. 4CONV, ResNet12, ResNet18 and WRN are the exploited FEM's architectures. "Dec” is the abbreviation of ”Decision". 
}
\begin{center}
\resizebox{\textwidth}{!}{
\begin{tabular}{lcccccc}
\toprule 
\multicolumn{1}{l}{\multirow{2}{*}{\textbf{Method}}}
& \multicolumn{1}{c}{\multirow{2}{*}{\textbf{Venue}}}
& \multicolumn{1}{c}{\multirow{2}{*}{\textbf{Backbone}}} 
& \multicolumn{2}{c}{\textbf{mini-ImageNet}} 
& \multicolumn{2}{c}{\textbf{tiered-ImageNet}} 
\\ 
\cmidrule(l){4-7}
\multicolumn{1}{c}{}
& \multicolumn{1}{c}{}
& \multicolumn{1}{c}{}                          
& \textbf{$5$-way $1$-shot} & \textbf{$5$-way $5$-shot}  & \textbf{$5$-way $1$-shot}  & \textbf{$5$-way $5$-shot} \\  
\midrule
ProtoNet  \cite{snell2017prototypical} & NeurIPS,2017 & 4CONV
& $49.42$   & $68.20$    & -                   & -    \\
MAML  \cite{finn2017model} & ICML,2018 & 4CONV
& $48.70$   & $63.11$    & -                   & -         \\
RelationNet  \cite{sung2018learning} & CVPR,2018 & ResNet18 
& $52.48$   & $69.83$    & -                   & -          \\
Baseline  \cite{chen2019closer} & ICLR,2019 & ResNet18
& $51.75$   & $74.27$    & -   & -          \\
Baseline++  \cite{chen2019closer} & ICLR,2019 & ResNet18
& $51.87$   & $75.68$    & -   & -          \\
LEO  \cite{rusu2019meta} & ICLR,2019 & WRN
& $61.76$   & $77.59$    & $66.33$    & $81.44$      \\
TPN  \cite{liu2019learning} & ICLR,2019 & 4CONV
& $52.78$    & $66.42$    & $55.74$   & $71.01$   \\

AM3  \cite{xing2019adaptive} & NeurIPS,2019 & ResNet12
& $65.30$   & $78.10$    & $69.08$    & $82.58$      \\
TapNet  \cite{yoon2019tapnet} & ICML,2019 & ResNet12
& $61.65$   & $76.36$    & $63.08$   & $80.26$          \\
CTM  \cite{li2019finding} & CVPR,2019 & ResNet18
& $64.12$   & $80.51$    & -                   & -          \\
DenseCls \cite{lifchitz2019dense} & CVPR,2019 & ResNet12 
& $62.53$   & $79.77$    & -                   & -          \\
MetaOpt  \cite{lee2019meta} & CVPR,2019 & ResNet12
& $62.64$   & $78.63$    & $65.99$    & $81.56$    \\
TEAM  \cite{qiao2019transductive} & ICCV,2019 & ResNet12
& $60.07$   & $75.90$    & -    & -    \\
DWC \cite{dvornik2019diversity} & ICCV,2019 & ResNet12
& $63.73$   & $81.19$    & $70.44$    & $85.43$    \\
S2M2  \cite{mangla2020charting} & WACV,2020  & WRN
& $64.93$   & $83.18$    & $73.71$    & $88.59$    \\
Fine-tuning  \cite{dhillon2020baseline} & ICLR,2020 & WRN
& $65.73$    & $78.40$    & $73.34$   & $85.50$   \\
DSN-MR  \cite{simon2020adaptive} & CVPR,2020 & ResNet12
& $64.60$    & $79.51$    & $67.39$   & $82.85$   \\
MABAS  \cite{kim2020model} & ECCV,2020 & ResNet12
& $64.21$    & $81.01$    & -   & -   \\
DivCoop \cite{dvornik2020selecting} & ECCV,2020 & ResNet12
& $64.14$    & $81.23$    & -   & -   \\
HGNN \cite{chen2021hierarchical} & TCSVT,2021 & 4CONV
& $60.03$   & $79.64$    & $64.32$    & $83.34$   \\
URT \cite{liu2021universal} & ICLR,2021 & ResNet12
& $\underline{72.23}$    & $83.35$    & $80.30$   & $88.63$   \\
DC  \cite{yang2021free} & ICLR,2021 & WRN
& ${68.57}$   & $82.88$    & $78.19$    & $\underline{89.90}$   \\
BOIL   \cite{oh2021boil} & ICLR,2021  & ResNet12
& $66.80$             & $79.26$            & $\underline{80.79}$            & $87.92$    \\
MELR   \cite{fei2021melr} & ICLR,2021  & ResNet12
& $67.40$    & $\underline{83.40}$    & $72.14$   & $87.01$    \\
ODE   \cite{xu2021learning} & CVPR,2021  & ResNet12
& $67.76$    & $82.71$    & $71.89$   & $85.96$    \\
\midrule
\textbf{MDFM} (2-Dec)  &\textbf{Ours} & ResNet12
& ${74.91}$  
& ${83.88}$  
& ${84.17}$
& ${89.95}$   \\
\textbf{MDFM} (4-Dec)  &\textbf{Ours} & ResNet12
& $\textbf{75.70}$  
& $\textbf{86.04}$  
& $\textbf{84.57}$
& $\textbf{90.94}$   \\




\bottomrule
\end{tabular}
}
\label{table: comparison_results_mini_tiered} 
\end{center}
\end{table*}

\begin{table*}[!t]
\caption{
The $5$-way supervised few-shot classification accuracies on CIFAR-FS and FC100 with $95\%$ confidence intervals over $600$ episodes. "Dec” is the abbreviation of ”Decision". 
}
\begin{center}
\resizebox{\textwidth}{!}{
\begin{tabular}{lcccccc}
\toprule 
\multicolumn{1}{l}{\multirow{2}{*}{\textbf{Method}}}
& \multicolumn{1}{c}{\multirow{2}{*}{\textbf{Venue}}}
& \multicolumn{1}{c}{\multirow{2}{*}{\textbf{Backbone}}} 
& \multicolumn{2}{c}{\textbf{CIFAR-FS}} 
& \multicolumn{2}{c}{\textbf{FC100}} 
\\ 
\cmidrule(l){4-7}
\multicolumn{1}{c}{}
& \multicolumn{1}{c}{} 
& \multicolumn{1}{c}{}                          
& \textbf{$5$-way $1$-shot} & \textbf{$5$-way $5$-shot}  & \textbf{$5$-way $1$-shot}  & \textbf{$5$-way $5$-shot} \\ 

\midrule
ProtoNet  \cite{snell2017prototypical} & NeurIPS,2017 & 4CONV
& $55.50$   & $72.00$    & $35.30$    & $48.600$    \\
MAML  \cite{finn2017model} & ICML,2018 & 4CONV
& $58.90$   & $71.50$    & -                   & -         \\
RelationNet  \cite{sung2018learning} & CVPR,2018 & 4CONV 
& $55.00$   & $69.30$    & -                   & -          \\
TADAM  \cite{oreshkin2018tadam} & NeurIPS,2018 & ResNet12
& -    & -    & $40.10$   & $56.10$   \\
DenseCls \cite{lifchitz2019dense} & CVPR,2019 & ResNet12 
& -    & -    & $42.04$   & $\underline{57.63}$   \\
MetaOpt  \cite{lee2019meta} & CVPR,2019 & ResNet12
& $72.00$   & $84.20$    & $41.10$    & $55.50$    \\
TEAM  \cite{qiao2019transductive} & ICCV,2019 & ResNet12
& $70.43$    & $81.25$   & -                   & -    \\
MABAS  \cite{kim2020model} & ECCV,2020 & ResNet12
& $73.24$    & $85.65$    & $41.74$   & $57.11$   \\
Fine-tuning  \cite{dhillon2020baseline} & ICLR,2020 & WRN
& $\underline{76.58}$    & ${85.79}$    & $\underline{43.16}$   & ${57.57}$   \\
DSN-MR \cite{simon2020adaptive} & CVPR,2020 & ResNet12
& $75.60$    & $\underline{86.20}$    & -   & -   \\

\midrule
\textbf{MDFM} (2-Dec)  &\textbf{Ours} & ResNet12
& ${81.68}$  
& ${88.57}$  
& ${46.97}$
& ${59.96}$   \\
\textbf{MDFM} (4-Dec)  &\textbf{Ours} & ResNet12
& $\textbf{82.20}$  
& $\textbf{89.75}$  
& $\textbf{48.69}$
& $\textbf{63.58}$   \\

\bottomrule
\end{tabular}
}
\label{table: comparison_results_cifar_fc100} 
\end{center}
\end{table*}

\begin{table*}[!t]
\caption{The $5$-way semi-supervised few-shot classification accuracies on mini-ImageNet and tiered-ImageNet with $95\%$ confidence intervals over $600$ episodes. "Dec” is the abbreviation of ”Decision". 
}
\begin{center}
\resizebox{\textwidth}{!}{
\begin{tabular}{lcccccc}
\toprule 
\multicolumn{1}{l}{\multirow{2}{*}{\textbf{Method}}}
& \multicolumn{1}{c}{\multirow{2}{*}{\textbf{Venue}}}
& \multicolumn{1}{c}{\multirow{2}{*}{\textbf{Backbone}}} 
& \multicolumn{2}{c}{\textbf{mini-ImageNet}} 
& \multicolumn{2}{c}{\textbf{tiered-ImageNet}} 
\\ 
\cmidrule(l){4-7}
\multicolumn{1}{c}{}
& \multicolumn{1}{c}{}
& \multicolumn{1}{c}{}                          
& \textbf{$5$-way $1$-shot} & \textbf{$5$-way $5$-shot}  & \textbf{$5$-way $1$-shot}  & \textbf{$5$-way $5$-shot} \\
\midrule
MSkM  \cite{ren2018meta} & ICLR,2018 & 4CONV
& $50.41$    & $63.39$    & $49.04$   & $62.96$   \\
TPN  \cite{liu2019learning} & ICLR,2019 & 4CONV
& $55.51$    & $69.86$    & $59.91$   & $73.30$   \\
LST \cite{li2019learning} & NeurIPS,2019 & ResNet12
& $70.10$    & $78.70$    & $77.70$   & $85.20$     \\
EPNet \cite{rodriguez2020embedding} & ECCV,2020  & ResNet12   
&  $\underline{75.36}$    &  $\underline{84.07}$    & $81.79$   & $88.45$   \\
TransMatch  \cite{yu2020transmatch} & CVPR,2020 & WRN
& $63.02$    & $81.19$    & -          & -         \\
ICI \cite{wang2020instance} & CVPR,2020  & ResNet12
& $71.41$             & $81.12$             & $\underline{85.44}$   & $\underline{89.12}$   \\

\midrule

\textbf{MDFM} (2-Dec)  &\textbf{Ours} & ResNet12
& $\textbf{80.45}$  
& ${87.65}$  
& $\textbf{87.82}$
& ${91.90}$   \\
\textbf{MDFM} (4-Dec)  &\textbf{Ours} & ResNet12
& ${80.42}$  
& $\textbf{88.09}$  
& ${87.72}$
& $\textbf{92.08}$   \\

\bottomrule
\end{tabular}
}
\label{table: semi_comparison_results_mini_tiered} 
\end{center}
\end{table*}

\begin{figure*}[t]
    \centering
    \subfigure[mini-ImageNet]
    {
        \includegraphics[width=0.45\linewidth]{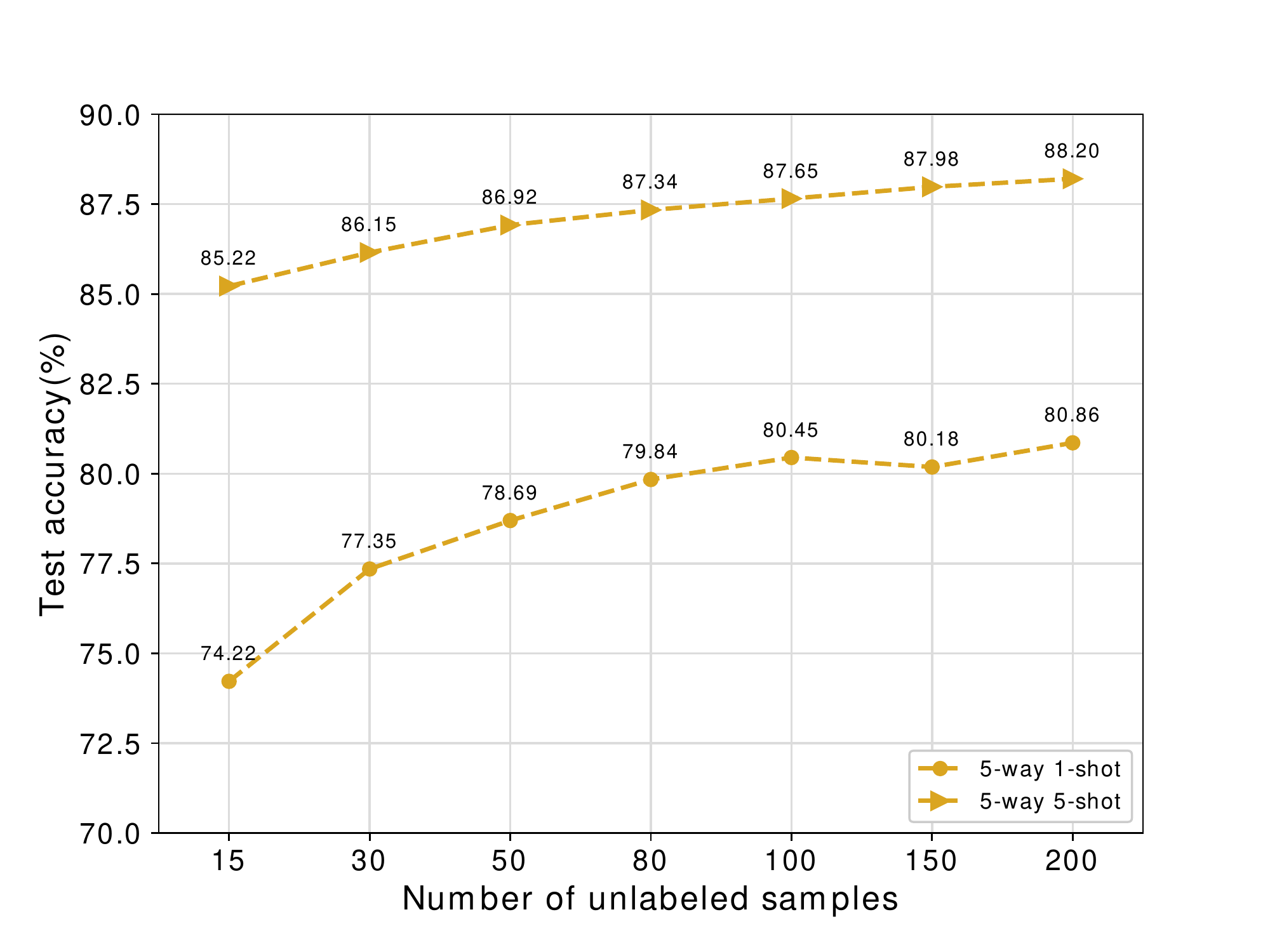}
    }
    \subfigure[tiered-ImageNet]
    {
        \includegraphics[width=0.45\linewidth]{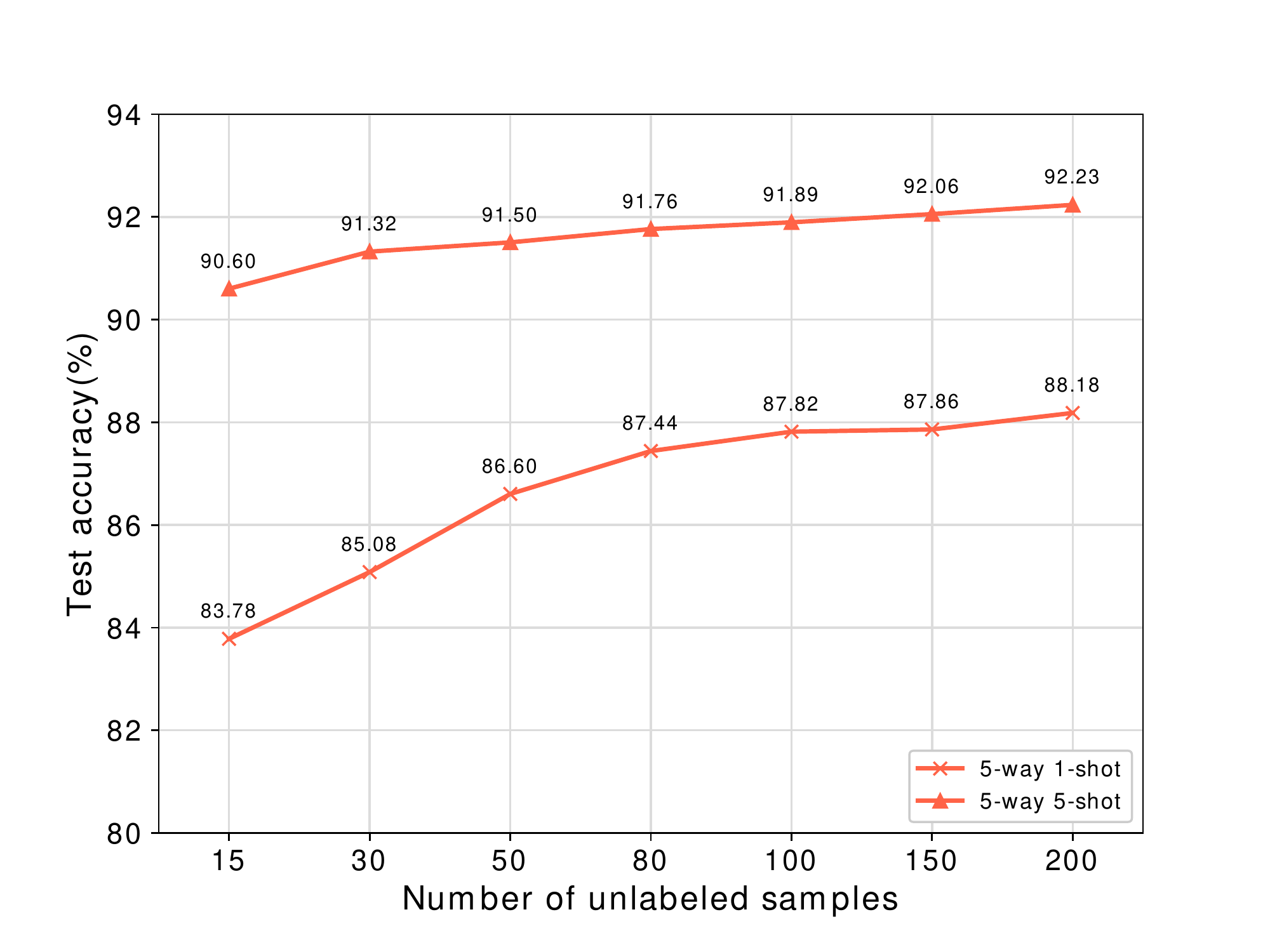}
    }
    \caption{The comparison results of semi-supervised few-shot classification with varied unlabeled samples on mini-ImageNet and tiered-ImageNet.}
    \label{fig: unlabel}
\end{figure*}

\subsection{Datasets}
We carry out experiments on five benchmark datasets, including mini-ImageNet \cite{vinyals2016matching}, tiered-ImageNet \cite{ren2018meta}, CIFAR-FS \cite{bertinetto2019metalearning}, FC100 \cite{oreshkin2018tadam}, and CUB \cite{wah2011caltech}. 
Both mini-ImageNet and tiered-ImageNet are the subsets of ImageNet dataset \cite{russakovsky2015imagenet}. mini-ImageNet consists of $100$ classes and tiered-ImageNet contains $608$ classes. For both datasets, the number of images for each class is $600$ and the size of each image is $84 \times 84$. We follow standard split as \cite{wang2020instance}, select $64$ classes as the base set, $16$ classes as the validation set and $20$ classes as the novel set for mini-ImageNet, and select $351$ classes as the base set, $97$ classes as the validation set and $160$ classes as the novel set for tiered-ImageNet. 
Both CIFAR-FS and FC100 are the subsets of CIFAR-100 dataset \cite{krizhevsky2009learning}, and consist of $100$ classes.
We follow the split introduced in \cite{bertinetto2019metalearning} to divide CIFAR-FS into $64$ classes as base set, $16$ classes as validation set, $20$ classes as novel set, and divide FC100 into $60$ classes as base set, $20$ classes as validation set, $20$ classes as novel set. All the image size is $32 \times 32$.
CUB totally includes $11,788$ images with $200$ categories. We follow the setting in ICI \cite{wang2020instance} to split it into $100$ classes as base set, $50$ classes as validation set and $50$ classes as novel set. The images are cropped into $84 \times 84$.

\begin{table*}[t]
\caption{Results of Multi-Decision Fusing with the supervised setting on mini-ImageNet on $5$-way $1$-shot case. "Dec" is the abbreviation of "Decision".}
\begin{center}
\resizebox{\textwidth}{!}{
\begin{tabular}{lccccccccccccccc}
\toprule
\textbf{Decisions} & $1$ & $2$ & $3$ & $4$ & $5$ & $6$ & $7$ & $8$ & $9$ & $10$ & $11$ & $12$ & $13$ & $14$ & $15$ \\
\midrule
Std-Dec         
& $\checkmark$ & & & & $\checkmark$ & $\checkmark$ & $\checkmark$ & & & & $\checkmark$ & $\checkmark$ & $\checkmark$ & & $\checkmark$ \\
Meta-Dec       
& & $\checkmark$ & & & $\checkmark$ & & & $\checkmark$ & $\checkmark$ & & $\checkmark$ & $\checkmark$ & & $\checkmark$ & $\checkmark$ \\
SS-R-Dec         
& & & $\checkmark$ & & & $\checkmark$ & & $\checkmark$ & & $\checkmark$ & $\checkmark$ & & $\checkmark$ & $\checkmark$ & $\checkmark$ \\
SS-M-Dec        
& & & & $\checkmark$ & & & $\checkmark$ & & $\checkmark$ & $\checkmark$ & & $\checkmark$ & $\checkmark$ & $\checkmark$ & $\checkmark$ \\
\midrule
ACC               & $68.2$ & $65.8$ & $72.3$ & $72.6$ & $71.7$ & $73.3$ & $73.5$ & $73.3$ & $73.4$ & $74.9$ & $74.1$ & $73.8$ & $75.2$ & $75.1$ &  $75.7$\\
\bottomrule
\end{tabular} 
}
\end{center}
\label{tab: Multi_Feature_Fusion}        
\end{table*}

\begin{figure*}[t]
    \centering
    \subfigure[mini-ImageNet]
    {
        \includegraphics[width=0.45\linewidth]{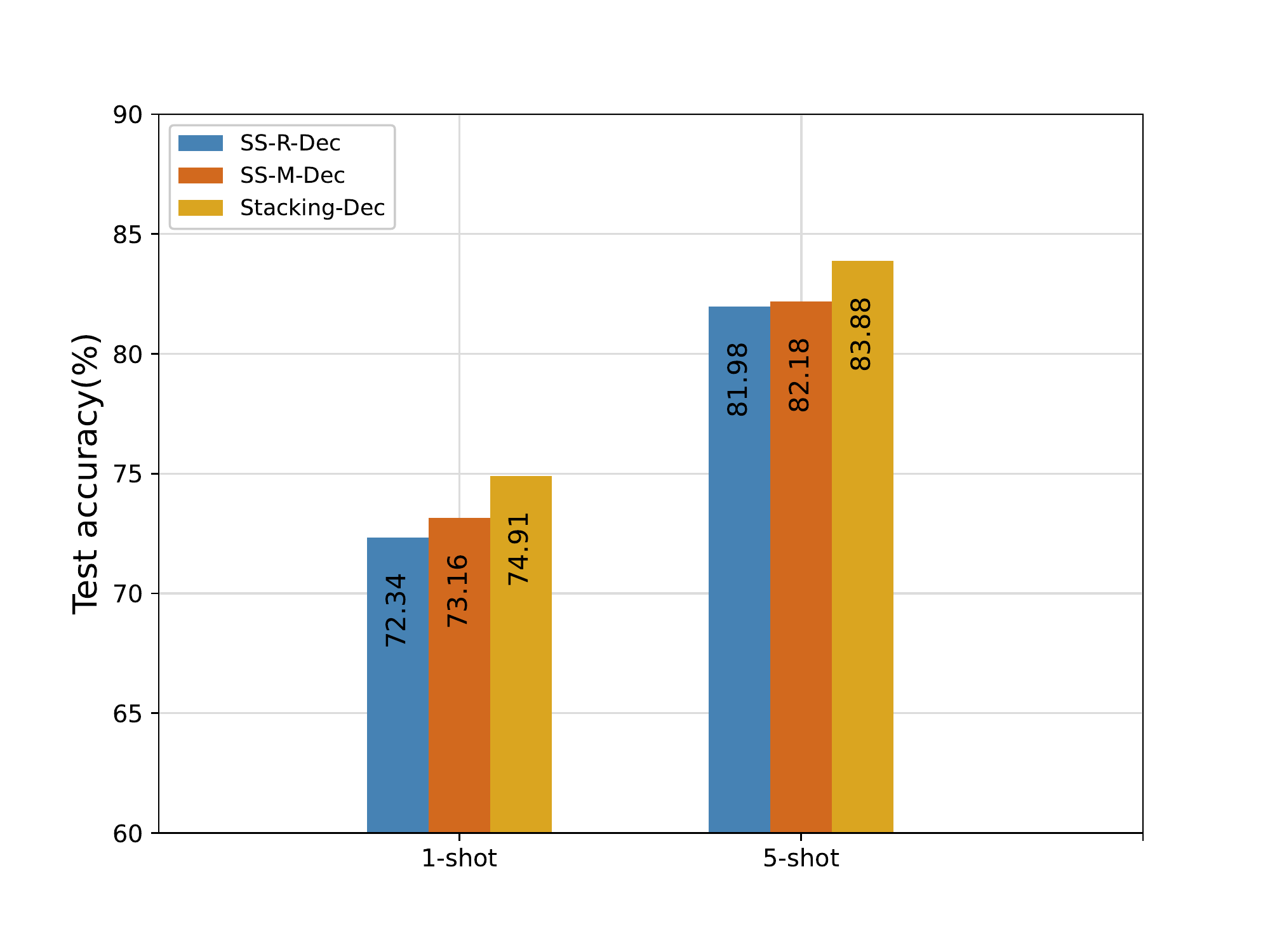}
    }
    \subfigure[tiered-ImageNet]
    {
        \includegraphics[width=0.45\linewidth]{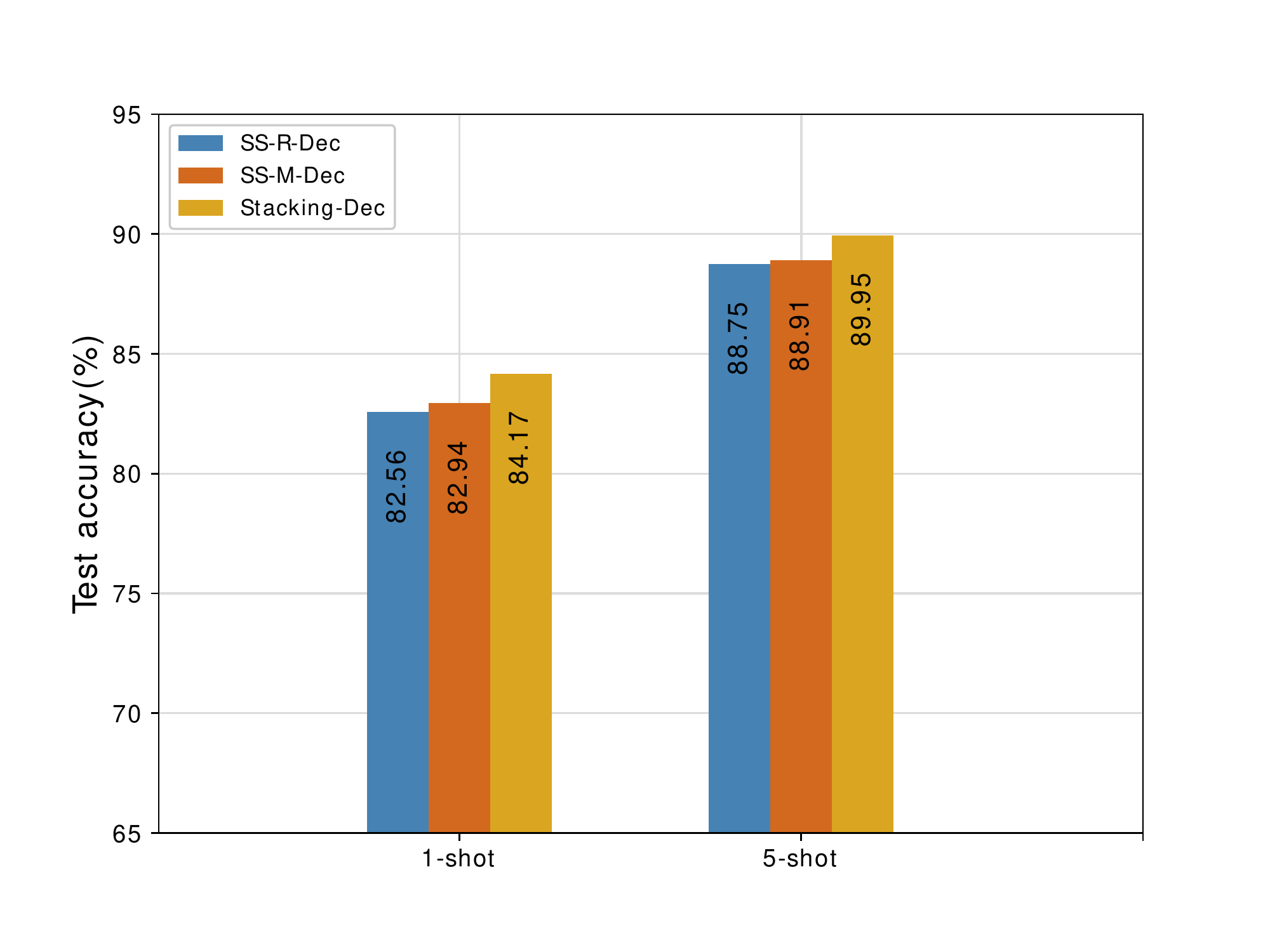}
    }
    \subfigure[CIFAR-FS]
    {
        \includegraphics[width=0.45\linewidth]{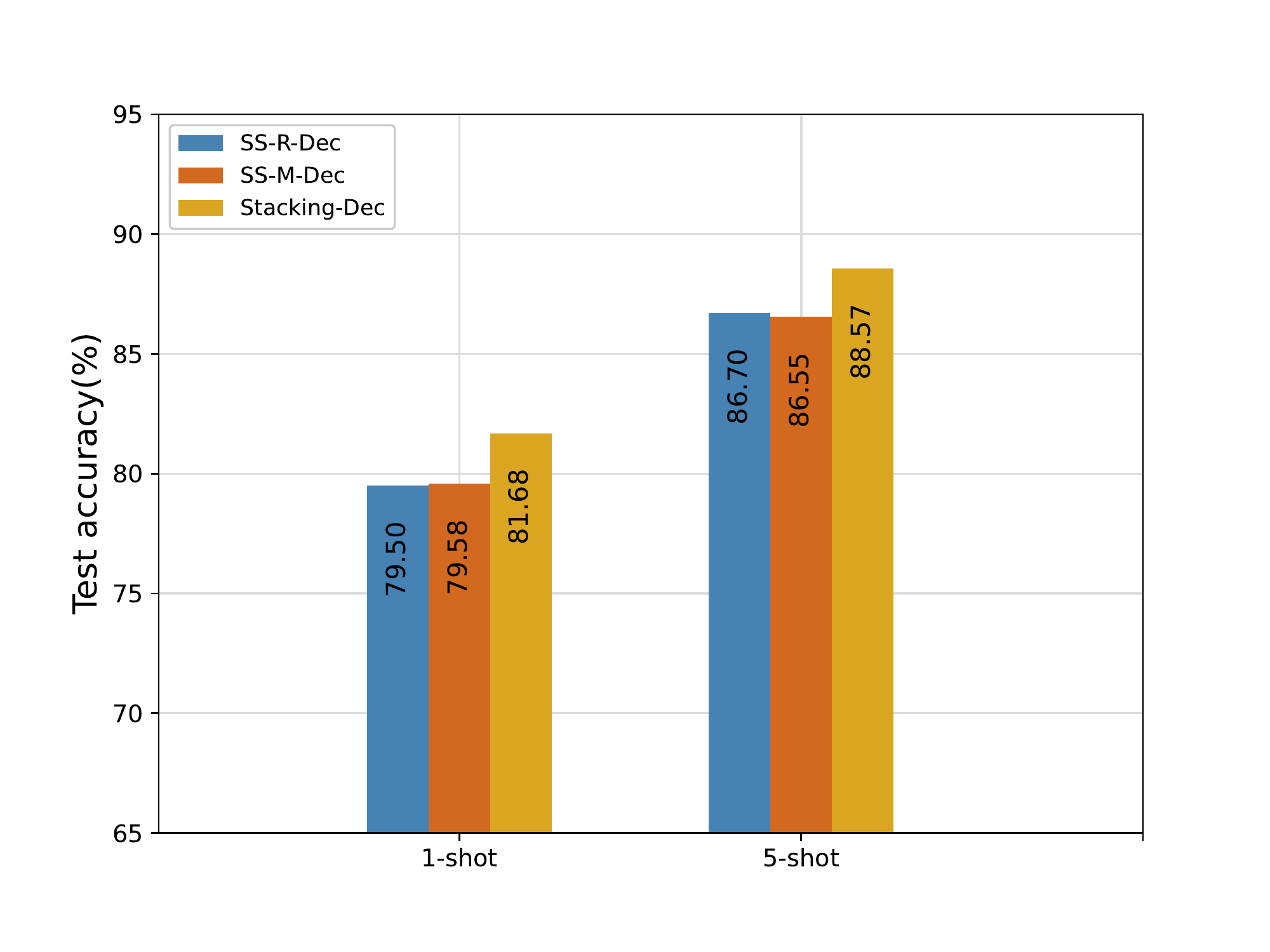}
    }
    \subfigure[FC100]
    {
        \includegraphics[width=0.45\linewidth]{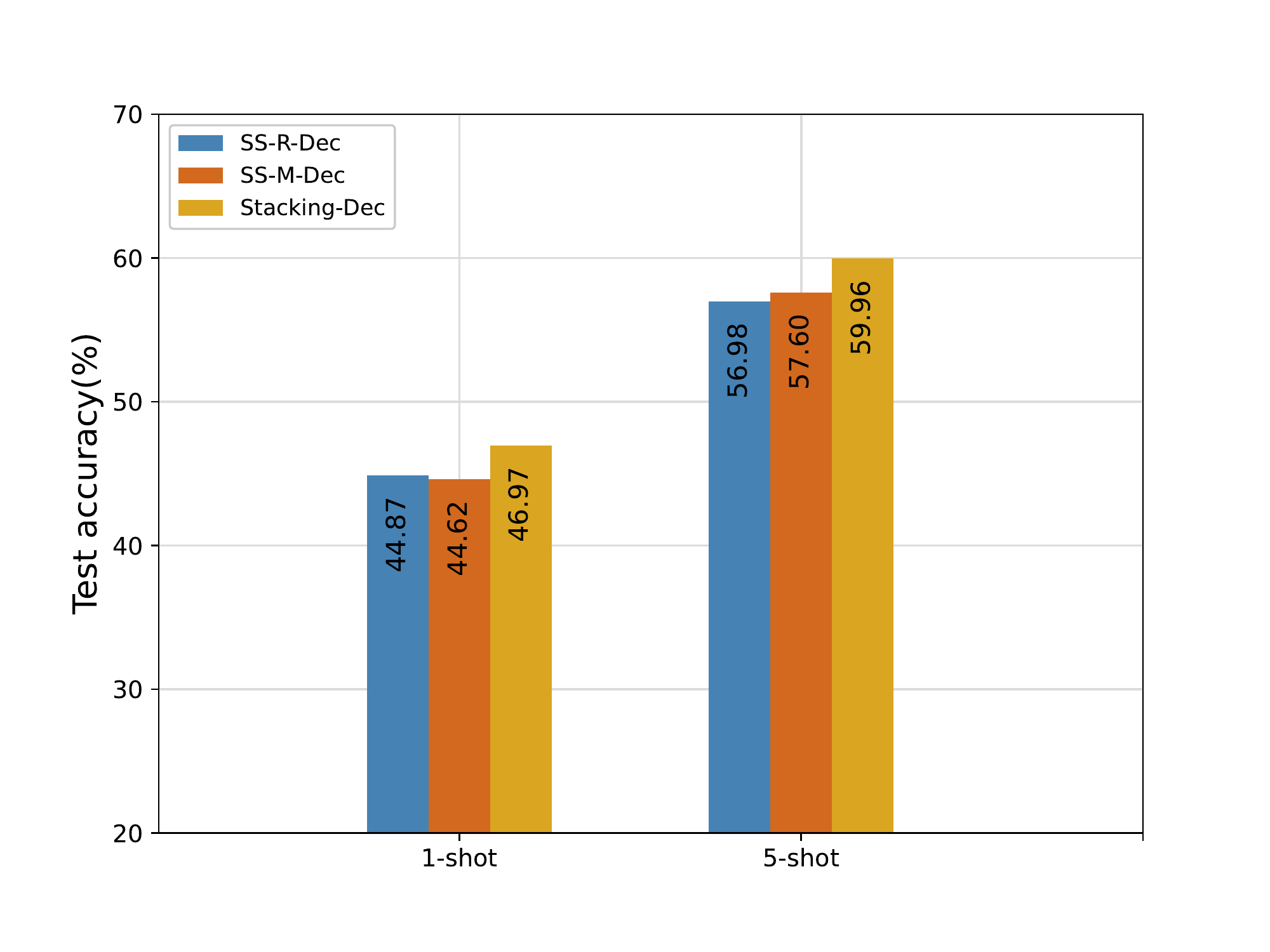}
    }
    \caption{Ablation studies to show the performances of different decisions on the supervised setting.}
    \label{fig: ablation study1}
\end{figure*}

\subsection{Implementation Details}
\label{subsec: Implementation Details}
In this paper, all the FEMs on different views adopt the ResNet12 \cite{he2016deep} backbone, consisting of four residual blocks ($3\times 3$ convolution layer, batch normalization layer, LeakyReLU layer), four $2\times 2$ max pooling layers, and four dropout layers. 
We adopt stochastic gradient descent (SGD) optimizer with Nesterov momentum ($0.9$) for the optimizer.
For the parameter $\eta$ in Equation (\ref{eqa: multi_view_classifier}), we fix it to $0.5$ for convenience.
We set the training epochs to $120$ and test over $600$ episodes with $15$ query samples per class for all the models.
Since that our method has decoupled the learning of representations and classifiers for the FSL, we have opportunities to deal with the extracted feature embeddings before classification. To this end, we introduce subspace transformation methods to strengthen the discrimination of the feature. Specifically, for the mini-ImageNet and tiered-ImageNet, we use Laplacian Eigenmap (LE) \cite{belkin2002laplacian}, and for the CIFAR-FS and FC100, we use the Principal Component Analysis (PCA) \cite{tipping1999probabilistic}.
Besides, we choose the Logistic Regression (LR) classifier with the default implementation of scikit-learn \cite{pedregosa2011scikit} and have no fine-tuning process when classifying the novel data.
For other settings, such as the data augmentation, the number of filters, we follow the ICI \cite{wang2020instance}.

\begin{figure*}[t]
	\begin{center}
		\includegraphics[width=0.8\linewidth]{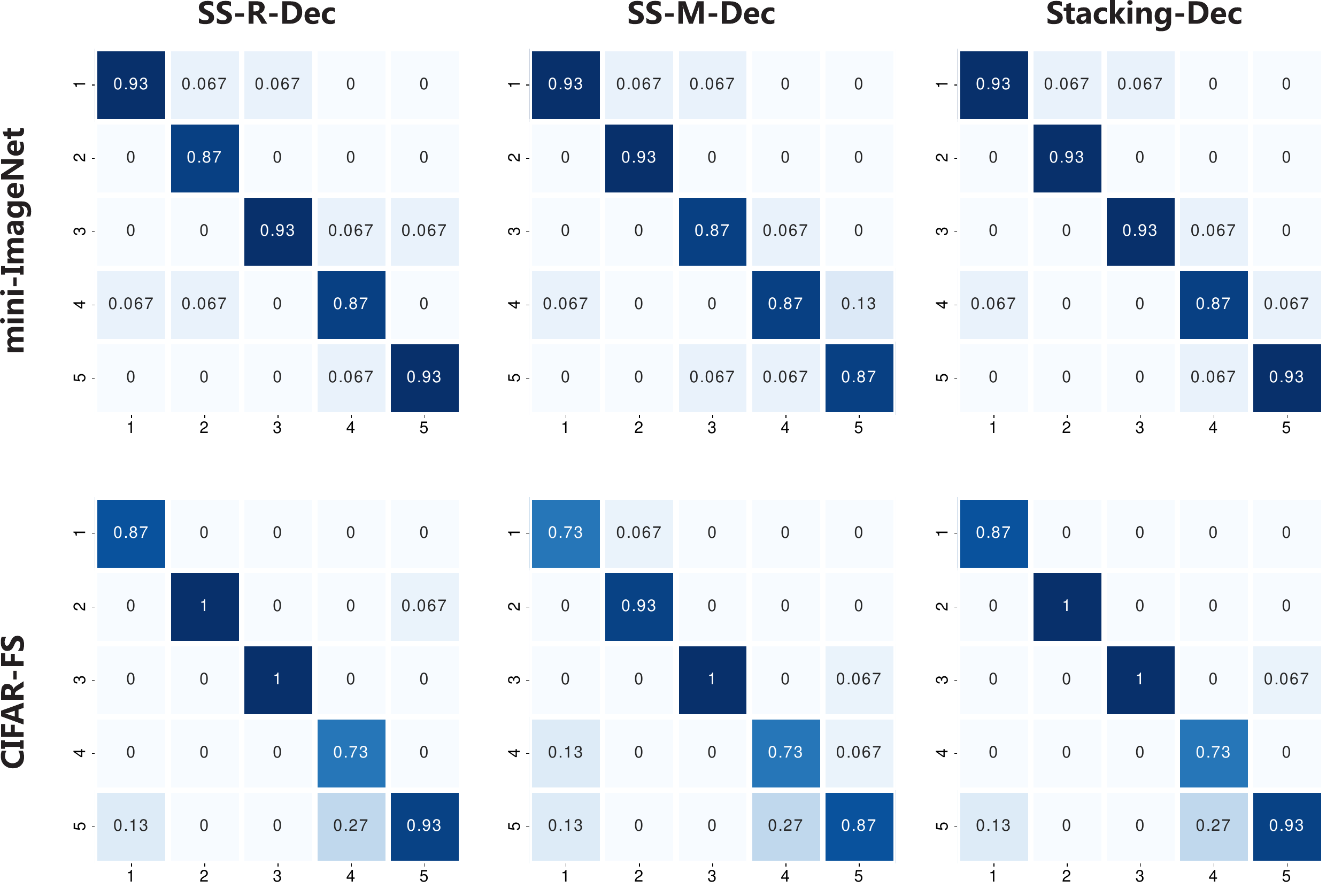}
	\end{center}
	\caption{Ablation studies to show the samples' confusion matrices of one episode on the supervised setting. }
	\label{fig: ablation_studie_2}
\end{figure*}

\begin{table}[h]
\begin{center}
\caption{Comparison results with fixed weights on $5$-way few-shot case. (a, b) denotes that the SS-R-Dec's weight is "a", and SS-M-Dec's weight is "b". Our method exploits the designed weighting mechanism to update the weights automatically for each episode. 
}
\setlength{\tabcolsep}{4.5mm}{
\begin{tabular}{lcccccc}
\toprule 
\multicolumn{1}{l}{\multirow{2}{*}{\textbf{Weight}}}
& \multicolumn{2}{c}{\textbf{mini-ImageNet}} 
& \multicolumn{2}{c}{\textbf{tiered-ImageNet}}
\\ 
\cmidrule(l){2-5}
\multicolumn{1}{c}{}
& \textbf{$1$-shot}   & \textbf{$5$-shot}  & \textbf{$1$-shot} & \textbf{$5$-shot}  \\  
\midrule
(0.1, 0.9)  
& {$\underline{73.9}$}             & $81.3$            & $83.5$           & $88.4$   \\
(0.3, 0.7)  
& $73.3$          & $83.4$            & $83.1$           & $89.2$   \\
(0.5, 0.5)  
& $73.2$            & {$\underline{83.6}$}          & 83.2           & {$\underline{89.7}$}   \\
(0.7, 0.3)  
& $73.4$           & $82.7$            & {$\underline{83.6}$}           & $89.2$   \\
(0.9, 0.1)  
&$72.7$      & $82.8$            & $83.1$           & $ 88.7$   \\
\midrule
\textbf{MDFM} 
& {$\textbf{74.9}$}   & {$\textbf{83.9}$}   & {$\textbf{84.2}$}  & {$\textbf{90.0}$}   \\
\bottomrule
\end{tabular}
}
\label{tab: fix_weight} 
\end{center}
\end{table}

\subsection{Experimental Results}
\label{subsec: Experimental Results}
We compare the proposed MDFM with several state-of-the-art methods, the results are listed in Table \ref{table: comparison_results_mini_tiered}, \ref{table: comparison_results_cifar_fc100}, \ref{table: semi_comparison_results_mini_tiered}. 
Here, we list some observations.

\textbf{i)} First, we look at the supervised results from Table \ref{table: comparison_results_mini_tiered}, \ref{table: comparison_results_cifar_fc100}. Obviously, our MDFM has far surpassed other approaches, especially on $5$-way $1$-shot case, at least 3.5\%, 3.8\%, 5.6\% and 5.5\% on mini-ImageNet, tiered-ImageNet, CIFAR-FS, FC100 datasets. The performances of our MDFM on the $5$-way $1$-shot case are even better than many other methods on the $5$-way $5$-shot case.
And on the $5$-way $5$-shot case, the MDFM also exceeds others at least 2.6\%, 1.0\%, 3.6\% and 6.0\% on mini-ImageNet, tiered-ImageNet, CIFAR-FS, FC100 datasets.

\textbf{ii)} Next, we compare our MDFM with other recently proposed multi-view based methods, including DenseCls \cite{lifchitz2019dense}, DWC \cite{dvornik2019diversity}, DivCoop \cite{dvornik2020selecting}, URT \cite{liu2021universal}. 
Obviously, our method outperforms them at least 3.5\% on the $5$-way $1$-shot case and at least 2.3\% on the $5$-way $5$-shot case.

\textbf{iii)} Then, we introduce a self-training strategy to extend our MDFM to the semi-supervised setting. 
Compared the results of MDFM in Table \ref{table: comparison_results_mini_tiered} and Table \ref{table: semi_comparison_results_mini_tiered}, we find that the unlabeled samples are really helpful to improve the performance for FSL.
Besides, from Table \ref{table: semi_comparison_results_mini_tiered}, compared our MDFM with other semi-supervised methods, our method (use $100$ unlabeled samples) also achieves excellent performances. 
We see that MDFM has significant improvements of at least 5.1\% and 4.0\% on mini-ImageNet with $5$-way $1$-shot and $5$-way $5$-shot case, 2.4\% and 3.0\% on tiered-ImageNet with $5$-way $1$-shot and $5$-way $5$-shot case.
In addition, for semi-supervised methods, the final results are influenced by the number of employed unlabeled samples. 
Thus, we use the mini-ImageNet and tiered-ImageNet as examples to observe the impact and list the results in Figure \ref{fig: unlabel}. The x-axis denotes the number of unlabeled samples. 
With the increase of unlabeled samples, the proposed method has became more effective. And the results start to saturate after $100$ unlabeled samples.

\textbf{iv)}
From Table \ref{table: comparison_results_mini_tiered}, \ref{table: comparison_results_cifar_fc100}, \ref{table: semi_comparison_results_mini_tiered}, we find that the 4-Dec based results are better than the 2-Dec based results in most cases, but in Table \ref{table: semi_comparison_results_mini_tiered} mini-ImageNet 5-way 1-shot case and tiered-ImageNet 5-way 1-shot case, the conclusion is inverse.
The reason is that introducing a large number of unlabeled data is helpful to calibrate the feature's distribution and weak the impact of our multi-decision fusion.

\subsection{Ablation Studies}
\label{subsec: Ablation Studies}

In this paper, we propose a multi-decision fusing method for few-shot learning. It is interesting to know the influence of \textit{fusing}. 
All the experiments are conducted on supervised setting.


\textbf{i)}
We list the results of only using one view's decision in Figure \ref{fig: ablation study1} and compare them with the fusing-view result (two view). From this figure, we can see that the performance of fusing-view improves significantly compared with the single-view, especially on the $1$-shot case. It has demonstrated the efficiency of our \textit{fusing} to some extent.

\textbf{ii)} To further evaluate the impact of the \textit{fusing}, we illustrate the experimental results of each class on mini-ImageNet and CIFAR-FS datasets.
Specifically, we randomly select one episode (include $5$ classes) on a $5$-shot case and show the corresponding confusion matrix of each view in {Figure \ref{fig: ablation_studie_2}}. Obviously, for a certain class, we obtain different results from different views, while the proposed MDFM at least achieves similar performance as the best result of single-view.
Thus, as the number of categories increases, the proposed method can naturally obtain more favorable results.



\textbf{iii)} As described in Section \ref{subsec: Discussion about Multiple Decisions}, the proposed method is capable of fusing multiple decisions. But the reported results in Table \ref{table: comparison_results_mini_tiered}, \ref{table: comparison_results_cifar_fc100}, \ref{table: semi_comparison_results_mini_tiered} only list the results of two ways (e.g., fuse 2-Dec and 4-Dec) for convenience. To further evaluate the proposed method, we carry out an experiment to show the performances of more kinds of fusing ways with the supervised setting on mini-ImageNet. 
The results are listed in {Table \ref{tab: Multi_Feature_Fusion}}.

\textbf{iv)} In theory, the proposed method can automatically assert weight to each decision through {Equation (\ref{eqa: omega_final})}. Thus, the ideal result is that: The more decisions are fused, the more choices are owned, and the better the results are obtained. 
From this table, we find that the conclusion is satisfactory. For example, $5$'s ACC is higher than $1$'s and $2$'s; $11$'s ACC is higher than $1$'s, $2$'s, $5$'s, $6$'s, and $8$'s; $15$'s ACC is higher than all the others.
Besides, we find that if the to-be-fused decisions have similar performances, the final fusing result may have significant improvement, such as ($1$, $2$, $5$), ($3$, $4$, $10$).

\textbf{v)} From {Table \ref{tab: Multi_Feature_Fusion}}, we find that the Std-Dec has outperformed many classical methods shown in {Table \ref{table: comparison_results_mini_tiered}}. There are two main reasons: 
On the one hand, the adopted standard FEM (Std-FEM) is captured from the ICI-based FEM \cite{wang2020instance}, which is a very strong FEM and enables the extracted features to have sufficient discrimination.
On the other hand, in the meta-test phase, we make process to the extracted feature embeddings, e.g., subspace transformation. The details are shown in {Section \ref{subsec: Implementation Details}}.
This operation can further enhance the discrimination of the data, which is very helpful for classification.

\textbf{vi)} In addition, it's not hard to find that with the fused views increase, the performance becomes saturated.
This is because the success of our approach depends largely on the diversity and complementarity of different views. As views increase, the more complete the model becomes, so the contribution of newly introduced views decreases.

\textbf{vii)} In order to reasonably integrate multi-view decisions, this paper proposes an weighting mechanism to dynamic weight different views of features, but how it works?
Here, we design an experiment to compare the results with fixed weights to our method, which is listed in {Table \ref{tab: fix_weight}}.
The results show that the updated weights are more reasonable for our method and the weighting mechanism is crucial. 

\begin{table}[!t]
\begin{center}
\caption{Comparison in cross-domain dataset scenario. Our MDFM is on supervised setting. $(\cdot)^\flat$ and $(\cdot)^\sharp$ indicate the reported results come from \cite{boudiaf2020transductive} and  \cite{mangla2020charting}. 
}
\setlength{\tabcolsep}{4.0mm}{
\begin{tabular}{lcccccc}
\toprule 
\multicolumn{1}{l}{\multirow{2}{*}{\textbf{Method}}}
& \multicolumn{2}{c}{\textbf{mini-ImageNet $\longrightarrow$ CUB}} 
\\ 
\cmidrule(l){2-3}
\multicolumn{1}{c}{}
& \textbf{$5$-way $1$-shot} & \textbf{5-way 5-shot}  \\  
\midrule
Baseline$^\flat$ \cite{chen2019closer}
& -                  & $53.1$    \\
MatchNet$^\flat$ \cite{vinyals2016matching}
& -                  & $53.1$    \\
MAML$^\flat$ \cite{finn2017model}
& -                  & $51.3$    \\
ProtoNet$^\flat$ \cite{snell2017prototypical}
& -                  & $62.0$    \\
RelationNet$^\flat$ \cite{sung2018learning}
& -                  & $57.7$    \\
GNN$^\flat$ \cite{tseng2020cross}
& -                  & $66.9$    \\
Neg-Cosine$^\flat$ \cite{liu2020negative}
& -                  & $67.0$    \\
LaplacianShot$^\flat$ \cite{ziko2020laplacian}
& -                  & $66.3$    \\
TIM-GD$^\flat$ \cite{boudiaf2020transductive}
& -                  & {$\underline{71.0}$}    \\
\midrule
MetaOpt$^\sharp$ \cite{bertinetto2019metalearning}
& $44.79$   & $64.98$    \\
Manifold Mixup$^\sharp$ \cite{verma2019manifold}
& $46.21$   & $66.03$    \\
S2M2$^\sharp$ \cite{mangla2020charting}
& {$\underline{48.24}$}   & $70.44$    \\
\midrule
\midrule
\textbf{MDFM}
& {$\textbf{60.56}$}    & {$\textbf{78.30}$}    \\
\bottomrule
\end{tabular}
}
\label{table: comparison_results_cross-domain} 
\end{center}
\end{table}

\subsection{{Cross-Domain Few-Shot Learning}}
\label{subsec: Cross-Domain}
After introducing multi-view information for selecting appropriate features for different categories, we believe that the MDFM is an extremely robust method in practical scenarios. To this end, we evaluate the proposed method with the supervised setting on a cross-domain dataset: e.g., mini-ImageNet $\longrightarrow$ CUB. The results are reported in {Table \ref{table: comparison_results_cross-domain}}. 
Obviously, compared to the state-of-the-art method, we have a significant improvement at least {12.3\%} on $1$-shot case and {7.3\%} on $5$-shot case.
Thus, the proposed MDFM would be powerful in real practice.

\section{{Conclusion}}

Few-shot learning (FSL) based tasks have a fundamental problem, e.g., distribution shift problem.
To address this challenge, we propose Multi-Decision Fusing Model (MDFM), which introduces multiple decisions to strengthen the FSL based model’s efficacy and robustness.
MDFM is a simple non-parametric method that can directly apply to the existing FEMs.
Experimental results have demonstrated the effectiveness of MDFM.
In our future work, it would be interesting to consider other fusing ways for FSL.


%



\section*{Acknowledgment}
The paper was supported by 
the National Natural Science Foundation of China (Grant No. 62072468), 
the Natural Science Foundation of Shandong Province, China (Grant No. ZR2019MF073), 
the Fundamental Research Funds for the Central Universities, China University of Petroleum (East China) (Grant No. 20CX05001A),
the Graduate Innovation Project of China University of Petroleum (East China) YCX2021117,
and the Graduate Innovation Project of China University of Petroleum (East China) YCX2021123.

\ifCLASSOPTIONcaptionsoff
 
\fi




\bibliographystyle{IEEEtran}
\bibliography{IEEEabrv.bib}
\end{document}